\documentclass[runningheads]{llncs}
\usepackage{graphicx}
\usepackage{amsmath,amssymb} 
\usepackage{color}

\newcommand{\comment}[1]{}

\newcommand{\ba}{\mathbf{a}}
\newcommand{\bb}{\mathbf{b}}

\newcommand{\be}{\mathbf{e}}

\newcommand{\bl}{\mathbf{l}}

\newcommand{\bt}{\mathbf{t}}
\newcommand{\bu}{\mathbf{u}}

\newcommand{\bx}{\mathbf{x}}

\newcommand{\bE}{\mathbf{E}}

\newcommand{\bI}{\mathbf{I}}

\newcommand{\bR}{\mathbf{R}}

\newcommand{\bS}{\mathbf{S}}

\newcommand{\bX}{\mathbf{X}}

\newcommand{\Pp}{\mathtt{P}}
\newcommand{\I}{\mathtt{I}}
\newcommand{\C}{\mathtt{C}}
\newcommand{\Rvar}{\tilde{{\bf R}}}

\newcommand\blfootnote[1]{%
  \begingroup
  \renewcommand\thefootnote{}\footnote{#1}%
  \addtocounter{footnote}{-1}%
  \endgroup
}

\begin{document}
\title{Stereo relative pose from line and point feature triplets} 

\titlerunning{Stereo relative pose from line and point feature triplets}
%
\author{Alexander Vakhitov\inst{1}
\and
Victor Lempitsky\inst{1}
\and
Yinqiang Zheng\inst{2}
}
%
\authorrunning{A. Vakhitov, V. Lempitsky and Y. Zheng}
%

\institute{Skoltech, Moscow, Nobelya Ulitsa 3, 121207, Russia \\ 
\email{\{a.vakhitov, lempitsky\}@skoltech.ru}\\
\and
NII, 2-1-2 Hitotsubashi, Chiyoda-ku, Tokyo 101-8430, Japan\\
\email{yqzheng@nii.ac.jp}}
\maketitle              

\blfootnote{The work is funded by the Russian MES grant RFMEFI61516X0003; a part of this work was finished when Alexander Vakhitov was visiting the National Institute of Informatics (NII), Japan, funded by the NII MOU/Non-MOU International Exchange Program.}
\begin{abstract}
Stereo relative pose problem lies at the core of stereo visual odometry systems that are used in many applications. In this work we present two minimal solvers for the stereo relative pose. We specifically consider the case when a minimal set consists of three point or line features and each of them has three known projections on two stereo cameras. We validate the importance of this formulation for practical purposes in our experiments with motion estimation. We then present a complete classification of minimal cases with three point or line correspondences each having three projections, and present two new solvers that can handle all such cases. We demonstrate a considerable effect from the integration of the new solvers into a visual SLAM system.  

\keywords{minimal solver, stereo visual odometry, generalized camera, relative pose, line features}
\end{abstract}

\section{Introduction}
Minimal solvers in computer vision are used to generate camera motion hypotheses from required minimal sets of feature correspondences, e.g. five feature point correspondences for single camera relative pose estimation~\cite{Nister2004PAMI}. Such solvers are mostly used as a source of motion hypotheses inside a RANSAC loop~\cite{fischler1987random}. They are useful in providing initialization for the optimization procedures at the core of state-of-the-art SLAM systems~\cite{cadena2016past}. For many pose estimation problems, such solvers have already been developed and are extensively used, e.g. to create large-scale structure from motion reconstructions involving thousands of images \cite{agarwal2011building}. It is important to develop minimal solvers taking line segment correspondences as input in addition to points. As recent works demonstrated~\cite{micusik2017structure,xu2017pose}, the use of line segment features can considerably improve accuracy and robustness of visual SLAM and structure from motion systems.



To the best of our knowledge, there is no minimal solver for stereo camera relative pose estimation which is efficient enough for real-time use  and does not rely on simplifying assumptions limiting its applicability. Thus, \cite{stewenius2005solutions} is computationally heavy for real-time use, \cite{pradeep2012egomotion} is non-minimal and \cite{Ventura2015} employs an approximate rotation model that is valid only for small rotations. 
In this work, we describe two solvers that aim to close this gap, giving an efficient minimal solution to the stereo camera egomotion from three feature triplets. We assume that there are two stereo cameras with projection matrices $\Pp_{1,1}=[\bI, \,\,\, \mathbf{0}],$ $\Pp_{1,2}=[\bI, \,\,\, \bb]$ for the first camera,  $\Pp_{2,1} = [\bR, \,\,\, \bt]$ and $\Pp_{2,2} = [\bR, \,\,\, \bb + \bt]$ for the second one, where the baseline $\bb$ is known. The goal of the solvers is to find $\bR$ and $\bt$. In each case, we use three feature triplets, where each triplet is a set of three\footnote{In the presence of two-view correspondences only, the overlapped stereo can be regarded as a non-overlapped stereo, and some solutions have been proposed as in \cite{Kneip2014,Ventura2015,stewenius2005solutions}. We exclude this case from consideration because the major focus of this work is on overlapped stereo systems.}  projections of a 3D line or a 3D point computed using $\{ \Pp_{\alpha}, \Pp_{\beta}, \Pp_{\gamma}\}$, $\alpha \neq \beta \neq \gamma.$

\begin{figure}[t!]
\centering
\includegraphics[width=0.4\textwidth,trim={2cm 20cm 7cm 0},clip]{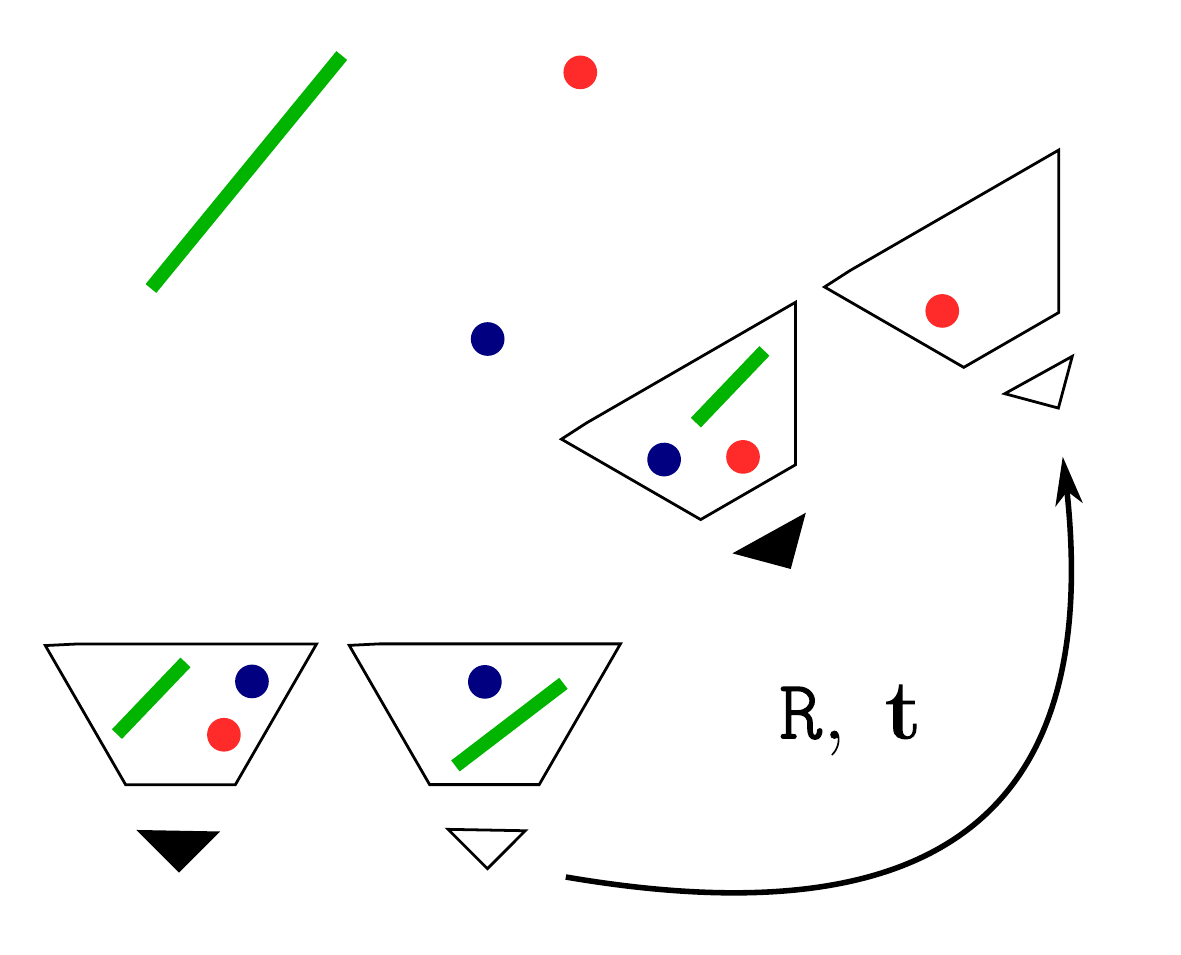}
\includegraphics[width=0.48\textwidth]{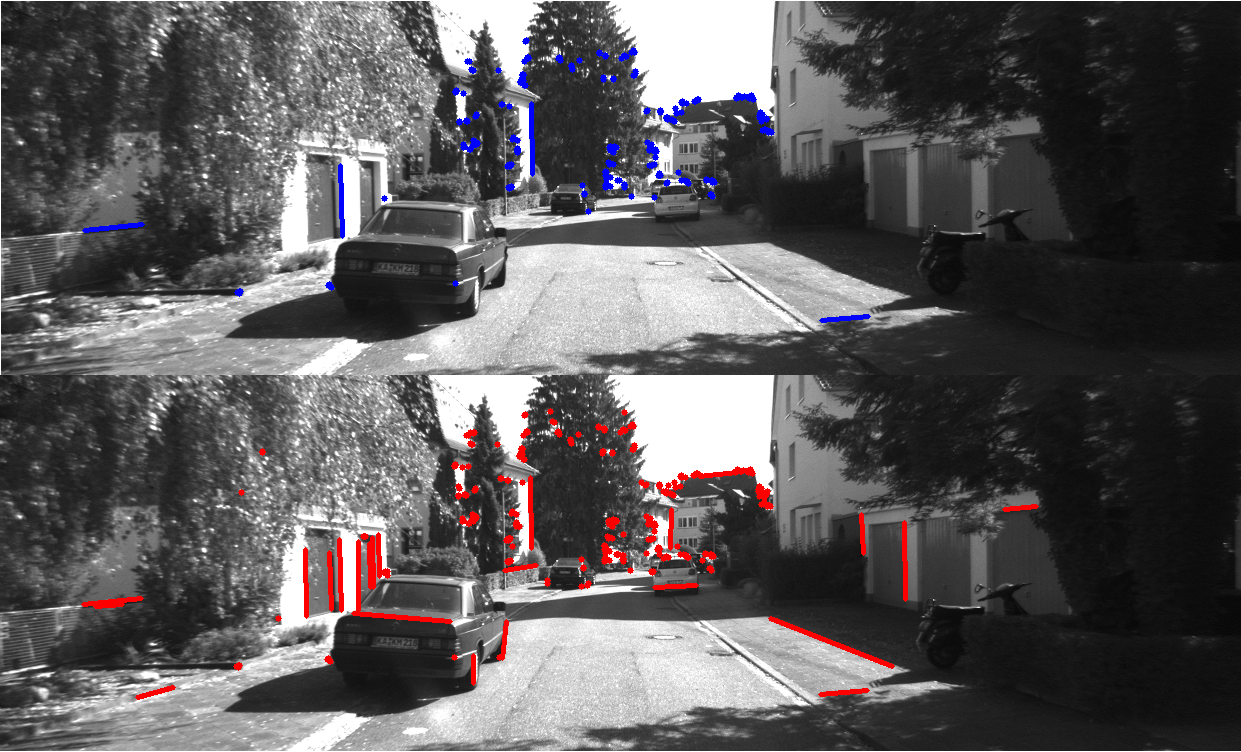}
\caption{Left:using three line or point features, each having exactly three projections, we seek to determine the relative pose of the two stereoviews. Right: the use of three-view matches (bottom) by the proposed solvers results in higher number of inliers compared to the use of four-view matches (top). We show the projections of the inlier correspondences on one of the images of KITTI sequence 0 chosen by the method Pradeep\cite{pradeep2012egomotion} using four-view matches (top) and by the proposed EpiSEgo solver using three-view matches (bottom).}
\end{figure}

While the ability to use features with three rather than four known projections may seem unnecessary for a stereo system, we show that such ability actually provides considerable benefits.
To illustrate this, we made a motivation experiment using the first sequence of the KITTI Odometry dataset \cite{Menze2015CVPR}. We use ORB~\cite{rublee2011orb} and LBD~\cite{zhang2013efficient} features and matched them between neighboring frames and across stereo-views. We then use the provided ground truth poses to estimate the ratio of inlier matches. We observe that for ORB matches the ratio of inlier matches across triplets of views is greater than those across quadruplets ($0.121$ vs $0.077$). For LBD line matches, the advantage is even greater ($0.019$ vs $0.005$). The advantage of relying on triplet matches is further corroborated in our experiments.
We develop two solvers covering any combinations of the point/line correspondences among the two view pairs. The first solver delivers 16 solutions, which is equal to the degree of the corresponding algebraic variety. It is impossible to obtain a solver for the formulated equations with the smaller number of solutions. The second solver outputs 32 solutions but is computationally simpler. Both are novel: to the best of our knowledge, no prior work describes a solution to the stereo camera relative pose problem for any combination of line/point features with three  projections or even only for the point features. 

Experiments show that our solvers are numerically stable and computationally efficient. More interestingly, by using point and line features simultaneously, our solvers work reliably for real scenarios. The use of three-view correspondences allows increasing the inlier cardinality and ratio, which not only facilitates the RANSAC procedure, but also reduces the risk of drifting in the case of long trajectories. 

To summarize, we make the following contributions. Firstly, we systematically explore the stereo ego-motion estimation problem in the case of a minimal set of three point and line features with three correspondences. Secondly, we develop new minimal solvers, which output a minimal number of solutions, and demonstrate the increase in accuracy and robustness of stereo egomotion estimation on simulated and real data. 

In Section \ref{sec:relatedworks}, we review the most closely related works on ego-motion estimation.  In Section \ref{sec:mathematicalformula}, we show the problem formulation and the complete categorization of the minimal point and line sets in any three views. We present the experiment results in Section \ref{sec:experiment}. 

\section{Related work}
\label{sec:relatedworks}


\paragraph{Non-overlapping fields of view:} To increase the coverage of the field of view (FoV) and to decrease the costs as much as possible, it became popular in recent years to use multiple cameras without overlapped FoVs. The generalized relative pose method proposed in~\cite{stewenius2005solutions} can be applied to estimate the relative pose of such multicamera systems, however it returns up to 64 solutions and is too computationally expensive for real-time use. 
To solve the problem in real-time, authors introduce certain approximations, e.g. Kneip and Li~\cite{Kneip2014} proposed to use non-minimal point sets and developed an approximated iterative optimization method, whose running speed is inappropriate for realtime applications. For acceleration, Ventura et al.~\cite{Ventura2015} linearized the rotation between two consecutive time frames, so the solver does not apply in the general visual odometry setting. 

\paragraph{Overlapping fields of view:} Binocular stereo systems with partially overlapping FoVs are preferable in terms of system calibration and metric reconstruction. 
To estimate the ego-motion of an overlapping stereo rig, Nister et al. \cite{Nister2004} proposed to use three points or two lines matched across all four views via triangulation. 
Chandraker~et~al.~\cite{Chandr2009} showed that the triangulation of four-view correspondences for ego-motion estimation is unstable, especially when the baseline is small. They proposed
instead to use three four-view line correspondences. Pradeep and Lim~\cite{pradeep2012egomotion} used assorted point and line features and developed several minimal solvers for any point and line combinations, as long as these features are simultaneously visible in all four views. 
Clipp et al.~\cite{clip2009} used point features in a mixed number of views, and Dunn et al.~\cite{dunn2011geometric} used similar input data and accelerated the solving speed by using the constraints in proper ways.  
Discarding the correspondences without projections onto both views of one stereo camera, one can use generalized absolute pose solvers~\cite{nister2004minimal,ramalingam2006generic,nister2007minimal,merzban2014simple,miraldo2015direct,kukelova2016efficient}. 
To summarize, no prior work addresses stereo relative pose problem for three features with three projections. Most of the studies consider the case of 4-view correspondences of only point features. 



\section{Stereo egomotion solvers}\label{sec:mathematicalformula}
\begin{table}
\centering
\begin{tabular}{l|c|c|c|c|c}
Case & Sect.  &  \multicolumn{4}{c}{Example} \\
 &  & \multicolumn{2}{c|}{1$^{st}$ cam } &    \multicolumn{2}{c}{2$^{nd}$ cam } \\
 \hline
 S3P & \ref{sect:same_main}  &  $a,b,c$  & $a,b,c$    &  $a$  & $b,c$   \\
 S2P1L & \ref{sect:same_main}  &  $a,b,\xi$  & $a,b,\xi$    & $a,\xi$   & $b$   \\
 S1P2L & \ref{sect:same_main}  &  $a,\xi,\theta$  & $a,\xi,\theta$    & $a$   & $\xi,\theta$   \\
 S3L & \ref{sect:line_only} &  $\xi,\theta,\gamma$  & $\xi,\theta,\gamma$    & $\xi,\theta$   & $\gamma$   \\
 S2L-1L & \ref{sect:line_only}  &  $\xi,\theta$  & $\xi,\theta,\gamma$    & $\xi,\theta,\gamma$   & $\gamma$   \\
 S2P-1L & \ref{sect:hard} &   $a,b$  & $a,b,\xi$    & $a,\xi$   & $b,\xi$   \\
 S1P1L-1P & \ref{sect:hard}  &  $a,\xi$  & $a,\xi,b$    & $a,\xi,b$   & $b$   \\
 S1P-2L & \ref{sect:hard}  &  $a,\xi$  & $a,\theta$    & $a,\xi,\theta$   & $\xi,\theta$   \\
 S1P1L-1L & \ref{sect:hard}  &  $a,\xi,\theta$  & $a,\xi$    & $a,\theta$   & $\xi,\theta$   \\
 S2P-1P & \ref{sect:hard}  &  $a,b,c$  & $a,b$    & $a,c$   & $b,c$   \\
 S1P-1P1L & \multicolumn{5}{c}{reduces to S1P1L-1P} \\
 S1P-2P & \multicolumn{5}{c}{reduces to S2P-1P} \\
\end{tabular}
\caption{The table enumerates all possible cases (excluding symmetries) and points to the section that discusses each case. Latin letters are for points and Greek are for lines (the details of the notation are discussed in the beginning of section \ref{sect:analysis}).}\label{tab:cat}
\end{table}
We address the problem of feature-based relative pose estimation for the binocular stereo camera, assuming that each line or point has exactly three projections.
The minimal set in this case consists of three features. 
Trifocal tensors provide a way to formulate constraints for the line and point features arising from three perspective views. Using the translation and rotation parameterizations (\ref{eq:t_expr}), (\ref{eq:rotparam}) which are  explained below, these trifocal constraints become third-order equations.
While for each line feature there are two such equations, for every point feature nine equations are obtained~\cite{hartley2003multiple}, of which only two are linearly independent. This effect complicates the solver construction. 

At the same time, in our problem formulation, for each feature there always exists a stereo camera such that the feature is projected onto both of its views (the \textit{main camera}). This simplifies the problem and allows to use projection constraints or two-view epipolar constraints between each view of the main camera and the view of the other camera. We use these approaches below and show that we can obtain 16 or 8 solutions using the proposed solvers, compared to 64 solutions using the solver \cite{stewenius2005solutions} for the same problem.

\subsection{Problem}
We assume that there are two binocular rectified and calibrated stereo cameras with the same known baseline. We are given a set of triplet feature correspondences. Each correspondence is a triplet. For point feature, a triplet is $(\bx_{i_1,\beta_1},\bx_{i_2,\beta_2},\bx_{i_3,\beta_3})$, where $\bx_{i,\beta}$ denotes a homogeneous vector of point projection's coordinates onto a view $\beta$ of a camera $i$. For a line feature, a triplet is $(\bl_{i_1,\beta_1},\bl_{i_2,\beta_2},\bl_{i_3,\beta_3})$ where $\bl_{i,\beta}$ denotes a vector of 2D line's coefficients of a 3D line's projection onto a view $\beta$ of a camera $i$.

W.l.o.g., we 
assume that the baseline has unit length ($\bb=[1.0.0]^T$) and the projection matrices $\Pp_{i,\beta}$ for a view $\beta$ of a camera $i$ are $\Pp_{1,1}=[\I, \,\,\, \mathbf{0}],$ $\Pp_{1,2}=[\I, \,\,\, \bb]$, $\Pp_{2,1} = [\bR, \,\,\, \bt]$ and $\Pp_{2,2} = [\bR, \,\,\, \bb + \bt]$. Our goal is to find $\bR,\bt$.

\subsection{Analysis of feature combinations}
\label{sect:analysis}
As long as there are exactly three projections for each feature, we use the following definition.

{\bf Definition:} {\it If a feature is projected onto both views of some stereo camera, this camera is called the {\bf main camera} for this feature.}

We use the following notation for feature/correspondence  combinations. We refer to problem as $S\alpha P \beta L - \gamma P \delta L$ when the first camera is the main for $\alpha$ points and $\beta$ lines, while the second one is the main for $\gamma$ points and $\delta$ lines.  To simplify the analysis, for those combinations having points we assume that the first camera is the main for at least one point feature. Some combinations are reducible to other ones by swapping the first and second cameras.

The categorization of the possible feature combinations is summarized in Tab.~\ref{tab:cat}. For a homogeneous minimal set, there are two possible feature divisions between the cameras: S2L-1L and S3L for lines, or S2P-1P and S3P for points. If we have two points and one line, we can get only S2P1L, S2P-1L, S1P1L-1P cases.
For one point and two lines, there are S1P2L, S1P1L-1L and S2L-1P cases. No other feature/correspondence combinations are possible.  

If all the features have the same main camera (i.e.\ S3L, S3P, S2P1L, S1L2P), they can be triangulated in the coordinate frame of this camera, and the problem reduces to generalized absolute pose \cite{nister2007minimal} for lines and points known to have 8 possible solutions. If a minimal set consists only of lines (S3L and S2L-1L), it admits a particular straightforward scheme of solution (``easy'' cases). 

The other situations  (S2P-1L, S1P1L-1P, S1P-2L, S1P1L-1L, S2P-1P) are the ``hard'' cases. They share two common properties: the features have different main cameras and there is at least one point in the feature set. Minimal solvers for them are the main contributions of the paper. 

In the next section, we propose two polynomial solver-based approaches for the ``hard'' cases. After that, we show how the other cases can be reduced to finding the roots of a single eight-degree polynomial, and then  a recently proposed method \cite{kukelova2016efficient} can be used. For the degeneracy analysis, see Supp. Mat. 

\subsection{``Hard'' cases} \label{sect:hard}
In this section, we consider the situation when the features have different main cameras and there is at least one point in the minimal set. 
Without loss of generality,
a camera is the first one if the first (and maybe the only) point feature is projected onto both views of this camera. We also assume that it is projected onto the first view of the second camera.
We use the first point to express the translation $\bt$ in terms of the point's depth and rotation matrix elements, as in \cite{stewenius2005solutions}. In particular, from an equation describing the point's projection onto the first view of the second camera we get
\begin{equation}\label{eq:t_expr}
\bt = \alpha \bu - \bR \bS,
\end{equation}
where $\bS$ is the point's position triangulated in its main camera's coordinates, $\bu$ is the homogeneous vector of the point's projection,  $\alpha$ is the depth constant. We will denote as $\bt_{\beta} = \delta_{\beta,2} \bb$ the translation of the view $\beta$ w.r.t. the stereo camera coordinate system, where $\delta_{i,j}=1$ iff $i=j$, else $\delta_{i,j}=0$.
We use the unit quaternion-based rotation parameterization:
{
\scriptsize
\begin{equation}\label{eq:rotparam}
\bR = 
\left[
\begin{array}{ccc}
a^2+b^2-c^2-d^2 & 2bc-2ad& 2bd+2ac \\
2bc+2ad& a^2-b^2+c^2-d^2& 2cd-2ab \\
2bd-2ac& 2cd+2ab& a^2-b^2-c^2+d^2 \\
\end{array}
\right],
\end{equation}
\begin{equation}\label{eq:unit_norm}
a^2+b^2+c^2+d^2 = 1.
\end{equation}
}
We have experimented with two ways of formulation of the polynomial equations for the stereo egomotion problem explained in the following paragraphs.
\subsubsection{Solver based on Epipolar/Pluecker constraints.}
We describe next a solver for the 'hard' cases which uses generalized epipolar constraints as in \cite{stewenius2005solutions}. If the feature is a point, we analyze the epipolar constraint arising from its projection onto the view $\beta$ of the first camera and onto the view $\gamma$ of the second camera. The epipolar line has the equation in  homogeneous coordinates $\bE_{1,\beta \to 2,\gamma} \bx_{1,\beta}$ using the essential matrix  $\bE_{1,\beta \to 2,\gamma}(\alpha, \bR) = [\be_{\beta,\gamma}]_{\times} \bR$ where
$\be_{\beta,\gamma} = \bt + \bt_{\gamma} \bb + \bR \bt_{\beta},$ 
$[\ba]_{\times}$ is a matrix of a cross product with a vector $\ba$.
Then, the point's projection lies on the epipolar line, which translates to the following constraint:
\begin{equation}\label{eq:point_epi}
    \bx_{2,\gamma}^T \bE_{1,\beta \to 2,\gamma} (\alpha, \bR) \, \bx_{1,\beta} = 0.
\end{equation}
For the point feature, we will get two constraints of the form (\ref{eq:point_epi}) with the unknowns $\bR$ and $\alpha$.
Using 3D line's projections onto the views of its main camera $j$ we compute a pair of 3D points lying on the line $\bX_1, \bX_2$. 
Assuming that $j=1$, we get the following expression for the line through projections of the points $\bX_1$ and $\bX_2$:
\begin{equation}
\lambda \bl_{i,\beta} = (\bR \bX_1 + \bt + \bt_{\beta}) \times (\bR \bX_2),
\end{equation}
where $\lambda$ is a scaling parameter. It leads to the following constraint:
\begin{equation}\label{eq:line_epi}
[\bl_{i,\beta}]_{\times} \bR \big( ( \bX_1 + \alpha \bu + \bt_{\beta}) \times  \bX_2 - \bX_2 \times \bt_{\beta} \big) = 0.
\end{equation}
Likewise, we obtain the following constraint for $j=2$:
\begin{equation}\label{eq:line_epi_2}
[\bl_{i,\beta}]_{\times} \bR^T \big( (\bX_1 -\alpha \bu) \times \bX_2 - \bX_2 \times \bt_{\beta} \big).
\end{equation}
A system of the constraints (\ref{eq:point_epi}), (\ref{eq:line_epi}) or (\ref{eq:line_epi_2}) can be formulated as
\begin{equation}\label{eq:epi_constr}
\mathtt{A} {\bf r} + \alpha \mathtt{B} {\bf r} = {\bf 0},
\end{equation}
where ${\bf r}$ is a vectorized matrix $\bR$, and $\mathtt{A}$ and $\mathtt{B}$ are coefficient matrices.

Substituting the parameterization (\ref{eq:rotparam}) into (\ref{eq:epi_constr}), we get four equations of degree three w.r.t.\ $a,b,c,d,\alpha$ and add to them the constraint (\ref{eq:unit_norm}). After formulating these equations over $\mathbb{Z}p$, we find using Maple~\cite{char2013maple} that the dimension of the quotient ring for the polynomial ideal is 32, see \cite{cox2006using} for details. 
Each term in the equations (\ref{eq:epi_constr}) after substitution of (\ref{eq:rotparam}) is of degree 2 w.r.t.\ $a,b,c,d$. We divide the equations by $a^2$, and denote $\tilde{b} = b/a$, $\tilde{c}=c/a,$ $\tilde{d}=d/a$. We choose $a$ as a divisor because it is close to one if the rotation is not big, which is the typical case for the SLAM systems. Finally, we get the constraints in the vector form:
\begin{equation}
\C(\tilde{b},\tilde{c},\tilde{d}) [1, \; \alpha]^T = {\bf 0},
\end{equation}
where $\C(\tilde{b},\tilde{c},\tilde{d})$ is a $4 \times 2$ matrix consisting of second-degree polynomials. All the $2 \times 2$ sub-matrices of $\C(\tilde{b},\tilde{c},\tilde{d})$ must have zero determinants. It gives six equations of degree four, which we multiply with all the monomials of $\tilde{b},\tilde{c},\tilde{d}$ of degree three and obtain 240 equations and then use them to construct an elimination template.

After the LU-decomposition of the template matrix, using the action monomial $\tilde{d}$ to construct an action matrix, we obtain the solutions by eigen-decomposition, find $\alpha$ from the null-space of $\C(\tilde{b},\tilde{c},\tilde{d})$, find $a$ using the unit-norm constraint (\ref{eq:unit_norm}) and $\bt$ using (\ref{eq:t_expr}).

\subsubsection{Solver based on point projection constraints.}
For the this solver, we apply the known preprocessing rotation $\Rvar$ to the projections of all the features to the views of the second camera. $\Rvar$ is chosen so that the first point's projection is in the image center: $\bu= [0,0,1]^T$, see (\ref{eq:t_expr}) for the definition of $\bu$. The baseline vectors of the cameras become different, we denote them as $\bb_j$, where $j=1,2$ is the stereo camera index, and get $\bb_1=\bb$ and $\bb_2 = \Rvar \bb$.

We define a function $\pi_{1,\beta}(\bR, \alpha, \bX)$ describing the point projection process, which takes a 3D point $\bX$ expressed in the first camera's coordinate frame and outputs the homogeneous point projection coordinates to view $\beta$ of the second camera:
\begin{equation}
\pi_{1,\beta}(\bR, \alpha, \bX) = \bR (\bX - \bS) + \alpha \bu + \bt_{\beta},
\end{equation}
which is a standard point projection equation after we substitute the translation according to (\ref{eq:t_expr}).
By noting that the rotation from the second to the first camera is $\bR^T$ and the translation is $-\bR^T \bt = -\alpha \bR^T \bu + \bS$, using (\ref{eq:t_expr}), we formulate a similar function $\pi_{2,\beta}(\bR,\alpha,\bX)$ returning a projection of a 3D point $\bX$ expressed in the second camera's coordinate frame to a view $\beta$ of a first camera:
\begin{equation}
\pi_{2,\beta}(\bR, \alpha, \bX) = \bR^T (\bX - \alpha \bu) + \bS + \bt_{\beta}.
\end{equation}

We assume that the camera $j$ is the main one for the feature, and that the feature also has a projection onto a view $\beta$ of a camera $i \neq j$. The constraint for the point feature is obtained from $\pi_{i,\beta}(\bR, \alpha, \bX) = \lambda_x \bx_{i,\beta} $ by expressing and substituting the depth parameter $\lambda_x$:
\begin{equation}\label{eq:pt_constraint_proj}
\pi_{i,\beta}^{(k)}(\bR, \alpha, \bX_p) - \bx_{i,\beta}^{(k)} \pi_{i,\beta}^{(3)}(\bR, \alpha, \bX_p) = 0, \;\;\; k = 1,2,
\end{equation}
where $k$ is the coordinate index of the feature projection, and $\bX_p$ is found by triangulation using the point's projections onto the main camera views. The constraint for the line feature is:
\begin{equation}\label{eq:ln_constraint_proj}
\bl_{i,\beta}^T \pi_{i,\beta}(\bR, \alpha, \bX_j) = 0,\;\;\; j = 1,2.
\end{equation}

Using these constraints and substituting the parameterization (\ref{eq:rotparam}), we get a system of four equations:
\begin{equation}\label{eq:cmat_full}
\mathtt{D}(a,b,c,d) [1, \; \alpha]^T = {\bf 0},
\end{equation}
where $\mathtt{D}$ is a matrix of second-degree polynomials.

Generating in $\mathbb{Z}p$ the systems for all the possible feature combinations  together with a constraint (\ref{eq:unit_norm}) and using Maple~\cite{char2013maple} we find that the quotient ring dimension and the number of solutions is 16.

From the system (\ref{eq:cmat_full}) by subtracting equations we obtain one or two (S2L-1P) linearly independent second-degree equations free of $\alpha$. As before, by computing determinants we get fourth-degree equations. The final system consists of six fourth degree equations (or five for SP-2L, because one of the determinants is identically zero), one (or two, for SP-2L) $\alpha$-free second-degree equations, and a quadratic constraint (\ref{eq:unit_norm}). This system also leads to 16 solutions. 

The basis of the remainder quotient ring as a vector space is not the same for different feature combinations. In particular, for the S1P1L-1P and S1P1L-1L cases there is one particular basis, and another one for the combinations S2P-1L, S2P-1P, S2L-1P (see Supp.Mat.).

We solve the obtained system by constructing an elimination template. Denote the second degree equation obtained after subtraction as $f_1=0$, the unit norm constraint as $f_2=0$, the other equations as $g_i=0$, $i=1..6$. We form an equation set $F$ from $f_1$ multiplied with $a^2$, $f_2$ multiplied with $ab,ac,b^2,bc,c^2$, and $f_1,f_2,g_i$ for $i=1..6$. We multiply every equation from $F$ by $a,b,c,d$, then by $a,b,c$, then by $a,b$, then by $a$, and add all the equations obtained after every multiplication operation to a set $G$ of cardinality 975. It allows us to express all the basis monomials times the action variable $a$. 
We use LU decomposition and get the action matrix of size $16 \times 16$. It is four times smaller than in the case of the Epipolar/Pluecker constraints, so the eigendecomposition can be performed faster, but template construction and LU decomposition will be slower. 

\subsection{Only line features} \label{sect:line_only}
The previous analysis is missing two situations: when all the features are lines or when they all have the same main camera. Next, we describe how both situations lead to a second-degree polynomial system in three unknowns, and therefore can be addressed by an already developed method for this type of geometric computer vision problems.

The coefficients of the 3D line's projection coincide with the direction of the normal to the plane through the camera center and the 3D line. If we observe the line from three views, we know normals of three different planes containing the line: ${\bf n}_1, {\bf n}_2, {\bf n}_3$, and their triple product is equal to zero: ${\bf n}_1 \cdot ({\bf n}_2 \times {\bf n}_3) = 0$. We are going to use this fact to formulate the constraints as follows:
\begin{equation}
\bl_{j,\gamma}^T \bR^T (\bl_{i,1} \times {\bf l}_{i,2}) = 0,
\end{equation}
for $i=1,j=2$ or $i=2,j=1$ depending on whether the main camera for the feature is the first or the second one, $\gamma=1,2$ is the view number. Three such constraints formulated using (\ref{eq:rotparam}) result in a second-degree system with 4 unknowns, the fourth equation being the unit norm constraint.

\subsection{Single main camera} \label{sect:same_main} If all the features have the same main camera, their 3D coordinates w.r.t.\ this camera can be computed, and then the problem becomes a particular case of the generalized absolute pose problem (gP3P). In the case of three point features several methods are available in this case, e.g. \cite{kukelova2016efficient}. To the best of our knowledge, there are no papers analyzing the generalized absolute pose problem for the mixed point/line minimal sets. 

We propose to use the earlier introduced constraints (\ref{eq:pt_constraint_proj},\ref{eq:ln_constraint_proj}) here as well. Without loss of generality, we assume that the first camera is the main one for all the features, so we use the constraints with $\pi_{1,\beta}$ for $\beta=1,2$. The depth $\alpha$ enters the system linearly. It can be expressed as a linear combination of terms involving other unknowns. This way we obtain a system of three equations w.r.t the unknown rotation matrix parameters. 

 In both cases, we transform a system with four unknowns into a system of three quadrics in three unknowns $\tilde{b}=b/a,\tilde{c}=c/a,\tilde{d}=d/a$ by the use of the constraint (\ref{eq:unit_norm}) to remove the zero-order terms and division by $a^2$.
 
The recent work \cite{kukelova2016efficient} provides a framework to which our problem fits well, proposing a way to reduce a problem of three quadrics intersection to root-finding of a single eighth-degree polynomial 
by using the hidden variable method to construct a single eight-order polynomial w.r.t.\ $b$, and we customize the method by adaptively choosing the variable to hide using the condition numbers (see Supp.Mat.).

To sum up, we have considered all possible feature and correspondence combinations up to symmetries. In the cases for three line features or when all features share the same main camera, the method~\cite{kukelova2016efficient} can be applied. The remaining cases are the most difficult and we have proposed two new polynomial solvers to cope with them. Next, we demonstrate the benefits of using the proposed solvers in synthetic experiments and on real data.

\section{Experiments}
\begin{figure}\label{fig:add_noise}
\includegraphics[width=0.95\textwidth]{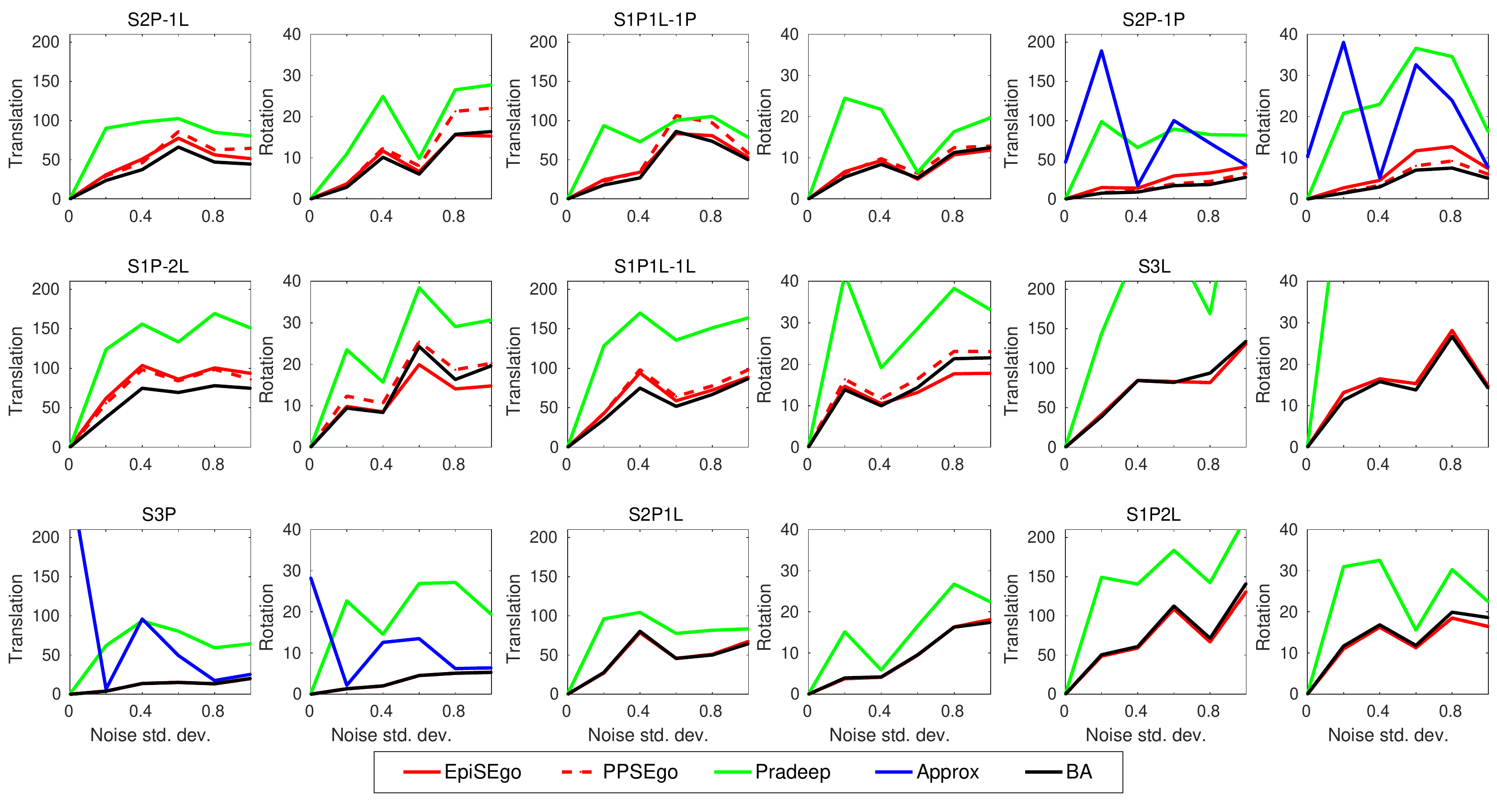}
\caption{The effect of additive noise variation on median relative translation and absolute rotation for each feature/correspondence combination. The accuracy of the methods degrades when noise is added. When fewer points are available, the translation error also grows. The new solvers PPSEgo and EpiSEgo are more accurate than the baselines and show almost the same accuracy as bundle adjustment (BA) started from the true solution and fitting to the noisy projections. The decrease in the error with growing noise happens simultaneously for BA and the proposed methods and can be explained by the non-linear nature of dependency between the projections and the SE3 transforms.}
\end{figure}
\label{sec:experiment}

\subsection{Simulated data}
\subsubsection{Setup:}
We perform a number of synthetic experiments to evaluate our method against the  non-minimal assorted features solver \cite{pradeep2012egomotion} ({\em Pradeep}). For point-only configurations, we also compare to an approximate minimal solver for generalized relative pose from points for small rotation \cite{Ventura2015} ({\em Approx}). Finally, we evaluate
bundle adjustment ({\em BA}) initialized with a true pose that uses the ''gold standard`` geometric feature reprojection error.
We use BA with oracle initialization as a reference to demonstrate the best realistically achievable accuracy. 
While our methods and Approx use minimal feature sets, Pradeep needs four projections for every feature. We have re-implemented Pradeep and used the original code of Approx. We evaluate both of the proposed solvers, namely the epipolar constraint-based (EpiSEgo) solver and the point projection constraint-based solver (PPSEgo). 


We assume that the stereo camera is rectified and the baseline is $\bb=[1;0;0]^T$. We also consider square images with the side of  1000 pixels and the vertical and horizontal view angle of 90$^{\circ}$. We fix the first stereo camera at the origin 
and randomly place the second camera. The points, as well as the 3D endpoints of the lines, are uniformly sampled from the box $B=[-1.5 , 2.5] \times [-1.5, 2.5] \times[12, 16]$. The distance between stereo cameras is sampled uniformly from the interval $[1,10]$. The second camera is rotated with angle uniformly sampled from the interval $[0,45]^{\circ}$ and around a random axis direction. If less than seven vertices of $B$ are visible, the pose is resampled. We add Gaussian noise with $\sigma=0.5$ pixels to the projections of the points and to line segments' endpoints. The lines have the length sampled uniformly from $[0.5,1.5]$, the line generating process follows \cite{vakhitov2016accurate}. Each experiment consists of 1000 random simulations for each of the possible feature/correspondence combinations. 

\subsubsection{Results.} We compute the median absolute rotation error (in degrees) and  relative translation error (in \%) for three overlapping sets of feature/correspondence combinations: 'hard' cases, 'easy' cases (i.e.\ three line features or features sharing the same main camera), and the point-only cases. To check numerical stability, we use zero additive noise and get the median (mean) rotation errors of $2 \times 10^{-9}$ ($5 \times 10^{-7}$) degrees for PPSEgo and $8 \times 10^{-7}$ ($2 \times 10^{-3}$) degrees for EpiSEgo, which is comparable to errors reported for similar solvers \cite{stewenius2005solutions}. PPSEgo is thus more numerically stable. 

The next experiment (Fig.~\ref{fig:add_noise}) shows that the difference diminishes when the noise is present.
Here we vary $\sigma$ from $0.0$ to $1.0$. The errors for all methods increase with the noise level, and the accuracy of the proposed solvers is close to the reference one of the BA and better than the one of Pradeep and Approx. The translation errors tend to be higher if more line features are in the minimal set. 

\begin{figure}\label{fig:var_rot}
\centering
\includegraphics[width=0.48\textwidth]{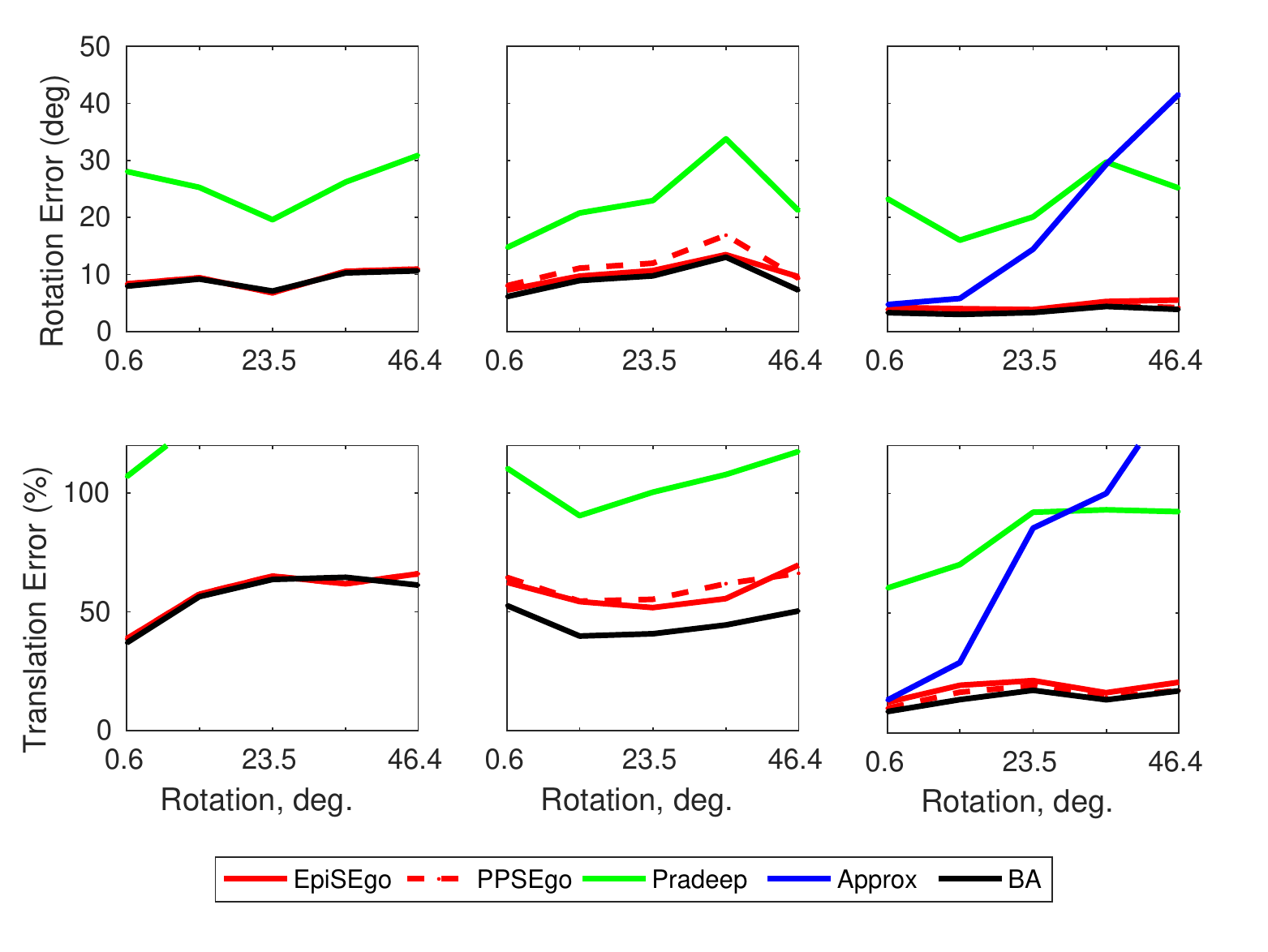}
\includegraphics[width=0.48\textwidth]{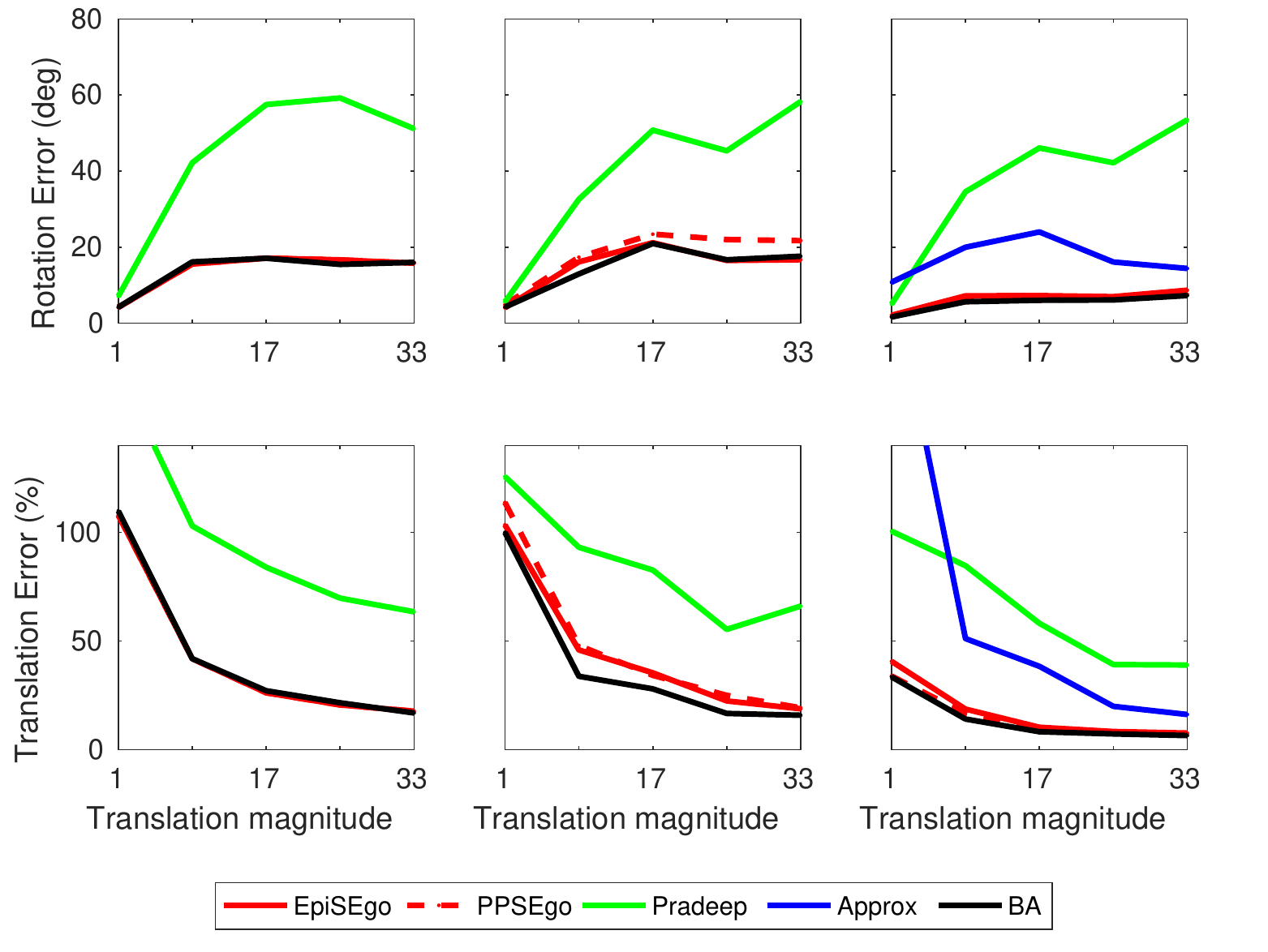}
\caption{The effects of rotation (left) and translation (right) magnitude variation on median relative translation and absolute rotation. The columns (left to right) correspond to: easy cases, hard cases, point-only combinations (S3P, S2P-1P). The new solvers PPSEgo and EpiSEgo have the lowest errors approaching the reference method (bundle adjustment with ground truth initialization).  
The new solvers PPSEgo and EpiSEgo are more accurate than the baselines (\textit{Pradeep} and \textit{Approx}, which is evaluated for point-only case).}
\end{figure}

Next, we vary the rotation magnitude from 0 to 45 degrees ( Fig.\ref{fig:var_rot}). The rotation accuracy for the SEgo methods approaches BA accuracy, while translation errors are bigger. The accuracy of both rotation and translation of Approx drops rapidly 
due to the use of small angle rotation approximation.
We then vary the translation magnitude from 1 to 33 ( Fig.~\ref{fig:var_rot}-right). Due to the choice of relative error to measure translation, we observe that translation accuracy increases, while the rotation errors grow and then stabilize. PPSEgo has slightly better translation and slightly worse rotation accuracy than EpiSEgo. Again, the accuracy of the new solvers approaches the reference (BA) and outperforms the baselines Pradeep and Approx.

\subsection{Real experiments}
\subsubsection{Matching between frames.} We use the processed and rectified grayscale stereo sequences of the KITTI dataset as input~\cite{Menze2015CVPR}. Given four views, we detect and match lines and points using the EDLines + LBD~\cite{akinlar2011edlines,zhang2013efficient} and ORB~\cite{rublee2011orb} algorithms implemented in OpenCV.

We evaluate one of our solvers {\bf EpiSEgo} against the baselines {\bf Pradeep}~\cite{pradeep2012egomotion}, 
{\bf Approx}~\cite{Ventura2015} and 
{\bf P3P}~\cite{gao2003complete}.
The Pradeep method takes as input four-view point and line correspondences. The P3P method emulates the classical approach of visual SLAM and takes the 
three-view 
correspondences constructed from both views of the first camera and the first view of the second camera. 
The Approx and our EpiSEgo methods work with three-view correspondences. While Approx uses only point features, our method employs both types of features.

To match point features between two images $I_l$ and $I_r$, for each feature from $I_l$  we find the closest one in $I_r$ by the descriptor distance. We reject the match if its reprojection error after triangulation is less than $\tau=5$ pixels. We match line segments in the same way, but without the reprojection validation. To find four-view correspondences, we match left and right images in both stereo pairs, and then match left images of the first and the second pairs. Denote the first stereo pair images as $I_l,I_r$, and the second stereo pair images as $I_l', I_r'$. We  find three view correspondences for each possible triplet of four images.

Then we run the classical RANSAC loop\cite{fischler1987random} with the threshold of $\tau$ pixels, $p=0.999$ and the initial outlier ratio of $0.5$. For the three-view correspondences, we triangulate a feature using its main camera, project onto the remaining view and compare the reprojection error to $\tau$. For the four-view correspondences, we choose one stereo pair, triangulate the feature using the projections onto its views, and then test the reprojection errors onto the views of the other stereo pair.

\begin{figure}\label{fig:real_me}
	\includegraphics[width=0.5\textwidth]{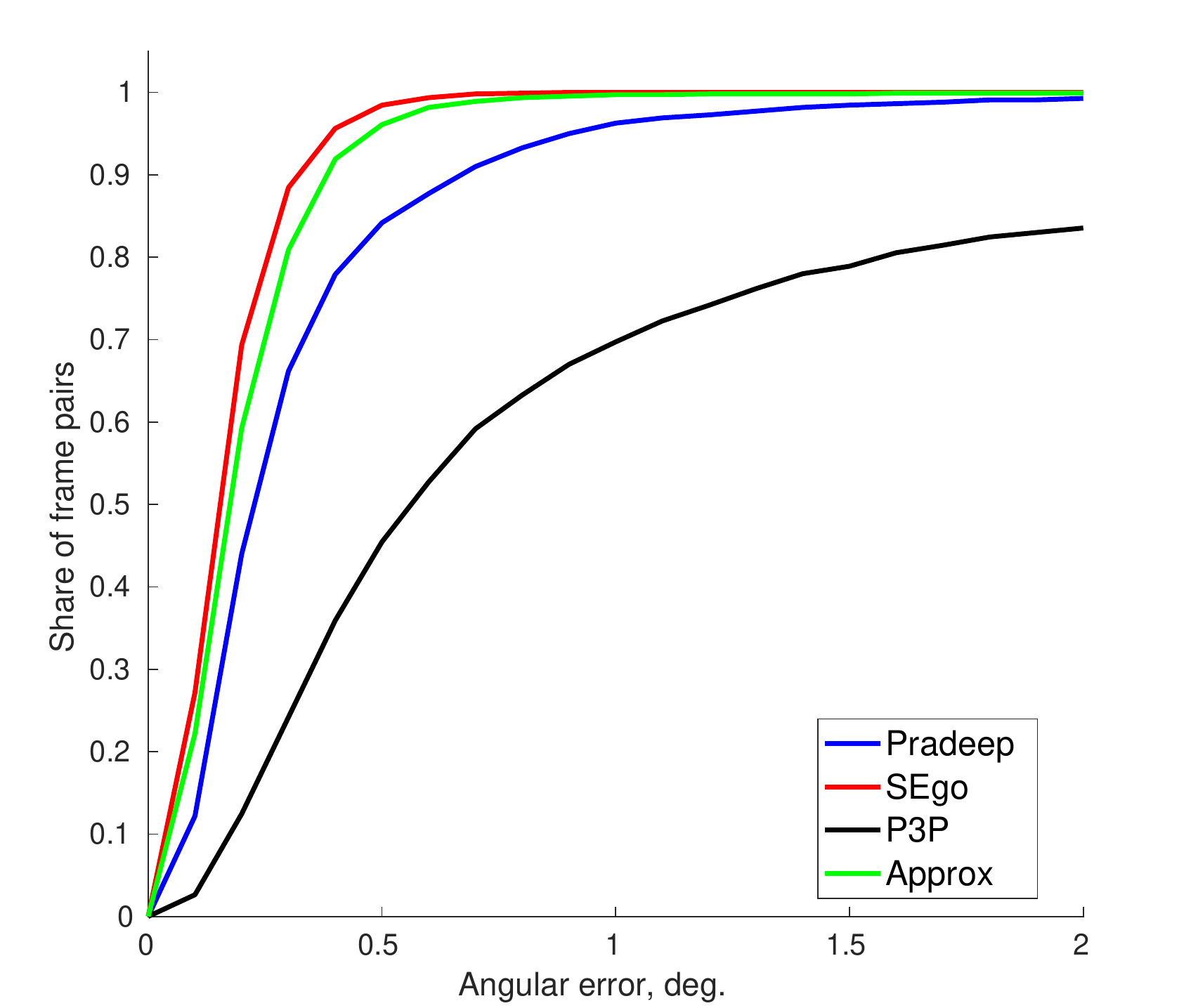}
	\includegraphics[width=0.5\textwidth]{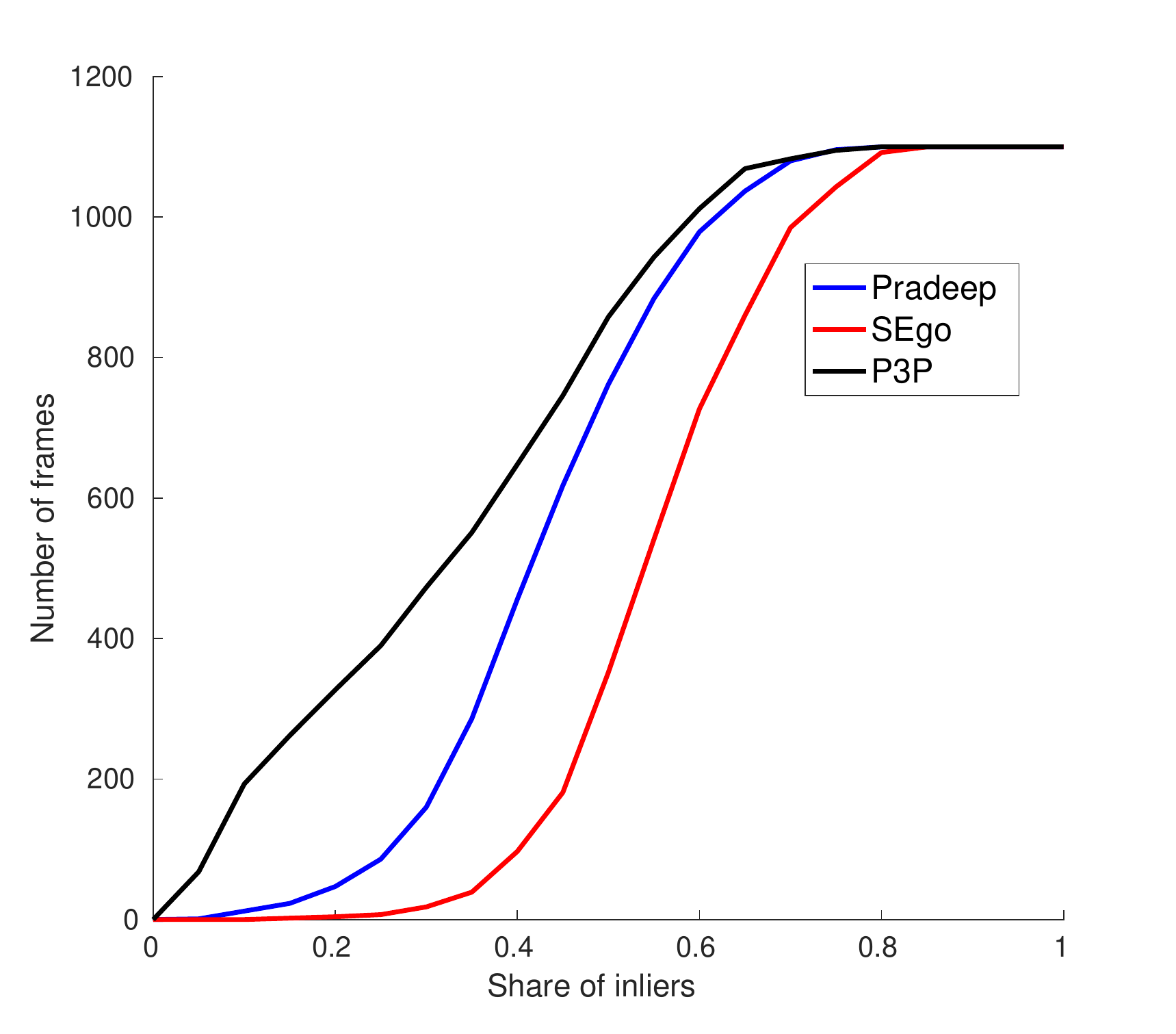}
	\caption{Results of the experiment on a KITTI sequence 6. Left: cumulative distribution for the rotation error in degrees. Right: cumulative distribution for the ratio of inliers. The proposed EpiSEgo using line and point feature triplets has higher accuracy compared to the baselines. It has higher inlier ratio than P3P and Pradeep. The use of all possible types of feature triplets rather than quadruplets (Pradeep) is beneficial in motion estimation. We also see a benefit from new solvers compared to the classical approach (P3P). Approx is excluded from the inlier plot because it relies only on point feature triplets.}
	\label{fig:ransaclm}
\end{figure}

In Fig.~\ref{fig:real_me} we show the results of the motion estimation experiment with the consecutive frame pairs of KITTI~\cite{geiger2012we} sequence 6. The use of three-view correspondences and point and line features helps the EpiSEgo to achieve high inlier ratios and lower pose estimation errors compared to the baselines. 

\begin{table}
\centering
\begin{tabular}{l|ccc|ccc}
Sequence & \multicolumn{3}{c|}{ORB-SLAM2} & \multicolumn{3}{c}{ORB-SLAM2+EpiSEgo} \\
&F\% & $t_{rel}$ & $r_{rel}$ &F\% & $t_{rel}$ & $r_{rel}$ \\
\hline
00 & 100\%&-&- & 0\%&62\%&$5*10^{-3}$ \\
 01 & 40\%&3\%&$10^{-4}$ & 40\%& 1.6\%&$10^{-4}$ \\
 02 & 100\%&-&-  & 0\% &53\%& $3.6*10^{-3}$ \\
 03 & 60\%&0.8\%&$10^{-5}$ & 0\% &3\%&$10^{-4}$\\
 04 & 40\% & 0.8\%&$10^{-5}$ & 0\% &0.8\%&$10^{-5}$\\
 05 & 100\%&-&- & 0\% &55\%&$4.8*10^{-3}$\\
 06 & 100\%&-&- & 0\%& 7\%&$10^{-4}$\\
 07 & 80\%&1.3\%&$10^{-4}$ & 0\%& 7\%&$10^{-4}$ \\
 08 & 100\%&-&- & 0\% &63\%&$4.6*10^{-3}$\\
 09 & 100\%&-&- &80\%& 63\%&$4.6*10^{-3}$\\
 10 & 100\%&-&- & 0\% &36\%&$2.8*10^{-3}$\\
\end{tabular}
\caption{The study of the robustness of the ORB-SLAM2 pipeline to the framerate decrease on the KITTI sequences 00-10. To make the task harder, we drop every second frame of each sequence. We compare the original and the modified pipeline that uses EpiSEgo solver for initialization in case of track loss. We report the percentage of runs where tracking was lost (F\%), the relative pose estimation errors as proposed by the dataset authors~\cite{geiger2012we}. After every second frame is dropped, there is the only sequence, which the original ORB-SLAM2 can track with probability more than 50\%. At the same time, the modified version shows radical improvement, as it tracks ten out of eleven sequences with probability more than 50\%. 
}\label{tab:tab2}
\end{table}
\subsubsection{Integration into visual SLAM pipeline.} While the previous experiment compares stereo egomotion methods at the task of relative pose estimation between stereo pairs, we also validate that such task can be used to improve modern stereo visual odometry pipelines. For this, we evaluate the system that integrates the proposed EpiSEgo solver into the ORB-SLAM2~\cite{mur2017orb} pipeline. 

The ORB-SLAM2 pipeline uses the previous frame pose as an initial guess to estimate the next frame pose within bundle adjustment. We modify it to run the EpiSEgo solver (point-only version) inside the RANSAC loop. 
The pose and the inliers estimated by RANSAC are used to initialize bundle adjustment. We run this algorithm each time the standard system loses the track. We do not include line features as they are absent in the original system.

While ORB-SLAM2 works well for the original sequences, it is important to study the robustness of the pipeline to the framerate decrease (which is equivalent to faster observer motion) which can happen in a real system. To do that, we drop {\it every second} frame of the sequence. Note that the uniform frame drop still enables the use of velocity-based pose prediction on which ORB-SLAM2 relies, provided the frames are separated by equal time periods. At the same time, it shows what can happen if motions become less predictable. Our experiments show that the ORB-SLAM2 often becomes unable to recover and loses the track, while the use of EpiSEgo solver can enable successful recovery from tracking losses. In Tab. \ref{tab:tab2}, we show the results of 5 runs for the original and modified ORB-SLAM2 on 0-10 KITTI sequences. We report the percentage of failures as well as relative rotation and translation errors proposed by the dataset authors. The modified version does not lose track with probability more than 50\% for all the sequences except the 9th, where a lack of tracked features in one moment is a possible problem. The original version is able to track with probability greater than 50\%  for only one sequence out of 11. The experiment shows that the integration of the stereo egomotion solver considerably increases the robustness of the system.

\section{Summary}
 \label{sec:conclusion}
 In this paper, we have proposed  new minimal solvers that can handle the stereo relative pose problem for any combinations of point and line three-view correspondences. This case was not addressed in the previous literature. We demonstrate that the problem is practical and leads to improved performance of a well-known SLAM system. 


\bibliographystyle{splncs}
\bibliography{egbib}

\pagebreak
\begin{center}
\textbf{\large Supplemental Materials: Stereo relative pose from line and point feature triplets}
\end{center}
\setcounter{equation}{0}
\setcounter{figure}{0}
\setcounter{table}{0}
\setcounter{page}{1}
\setcounter{section}{1}
In these Supplemental Materials we report additional theoretical material and experimental results not included into the main paper. We start with theory, then proceed to real and synthetic experiments. We attach the code reproducing the synthetic experiments. The  code will be available for the public at the time of publication.

\subsection{Method details}
\subsubsection{Polynomial solvers for the 'hard' cases (sect. 3.3)}
\paragraph{Epipolar/Pluecker constraints} The EpiSEgo solver is based on Epipolar/Pluecker constraints (5), (7-8). The construction of the polynomial system is described in sect. 3.3. It  involves three variables $\tilde{b},\tilde{c},\tilde{d}$. The quotient ring defined by the polynomial ideal has 32 solutions. It has a following basis for all the feature/correspondence combinations:
$$
[1,\tilde{b},\tilde{b}^2,\tilde{b}^3,\tilde{c},\tilde{c}^2,\tilde{c}^3,\tilde{b}\tilde{c},\tilde{b}\tilde{c}^2,\tilde{b}^2\tilde{c},\tilde{d},\tilde{d}^2,\tilde{d}^3,\tilde{d}^4,\tilde{d}^5,\tilde{b}\tilde{d},\tilde{b}\tilde{d}^2,$$
$$\tilde{b}\tilde{d}^3,\tilde{b}\tilde{d}^4,\tilde{b}^2\tilde{d},\tilde{b}^2\tilde{d}^2, \tilde{c}\tilde{d}, \tilde{c}\tilde{d}^2, \tilde{c}\tilde{d}^3, \tilde{c}\tilde{d}^4, \tilde{c}^2\tilde{d}, \tilde{c}^2\tilde{d}^2, \tilde{c}^3\tilde{d},$$
$$\tilde{b}\tilde{c}\tilde{d}, \tilde{b}\tilde{c}\tilde{d}^2, \tilde{b}\tilde{c}^2\tilde{d}, \tilde{b}^2\tilde{c}\tilde{d}].
$$
\paragraph{Point projection constraints} 
The PPSEgo solver is based on the point projection constraints (13-14). The polynomial polynomial system is constructed as described in the sect. 3.3. As verified by Maple\cite{char2013maple}, it has 16 solutions. The quotient ring defined by the polynomial ideal has a specific basis for a particular feature/correspondence combination. For the S1P1L-1P and S1P1L-1L cases it is
$$
[1,a,a^2,a^3,b,b^2,ab,a^2b,c,ac,a^2c,bc,d,ad,a^2d,bd],
$$
while for the remaining 'hard' cases S2P-1L, S2L-1P and S2P-1P it is
$$
[1,a,a^2,a^3,a^4,b,b^2,ab,a^2b,c,ac,a^2c,bc,d,ad,a^2d].
$$
\subsubsection{Condition number check for quadric intersection (sect. 3.4, 3.5)}
To solve the easy cases (sect. 3.4, 3.5) we use the technique described in \cite{kukelova2016efficient}. We have a system with 3 unknowns $\tilde{b}, \tilde{c}, \tilde{d}.$ We try to hide each of the variables. Then we compute the condition number of the system matrix $A$ (see formula (3) in the cited paper). We choose a matrix $A$ with a condition number closest to one by absolute value. We use for hiding the corresponding variable. 

Both PPSEgo and EpiSEgo use quadric intersection for the easy cases. During synthetic experiments we use condition number check in EPiSEgo but omit it in PPSEgo. We do so only to demonstrate the importance of this additional step in some situations. In real experiments, the EpiSEgo uses the condition number check. 

\subsubsection{Degenerate configurations}
Ambiguous configurations differ in 3D geometry while having identical projections of features. For the ambiguous configurations, the constraints become degenerate and the pose cannot be uniquely reconstructed. There are 3D feature configurations which constitute ambiguous configurations with anyhow placed cameras, such as three points on a line for the P3P problem, three parallel lines for the P3L problem or the features belonging to the same 3D line for the P1P2L and P2P1L problems. These configurations obviously remain ambiguous for all the problems considered in the paper. The analysis of all possible ambiguities/degeneracies in the set of all the possible combinations of feature/camera configurations is the part of future work.

\subsection{Real experiments}
\begin{figure}
\centering
\includegraphics[width=0.22\textwidth]{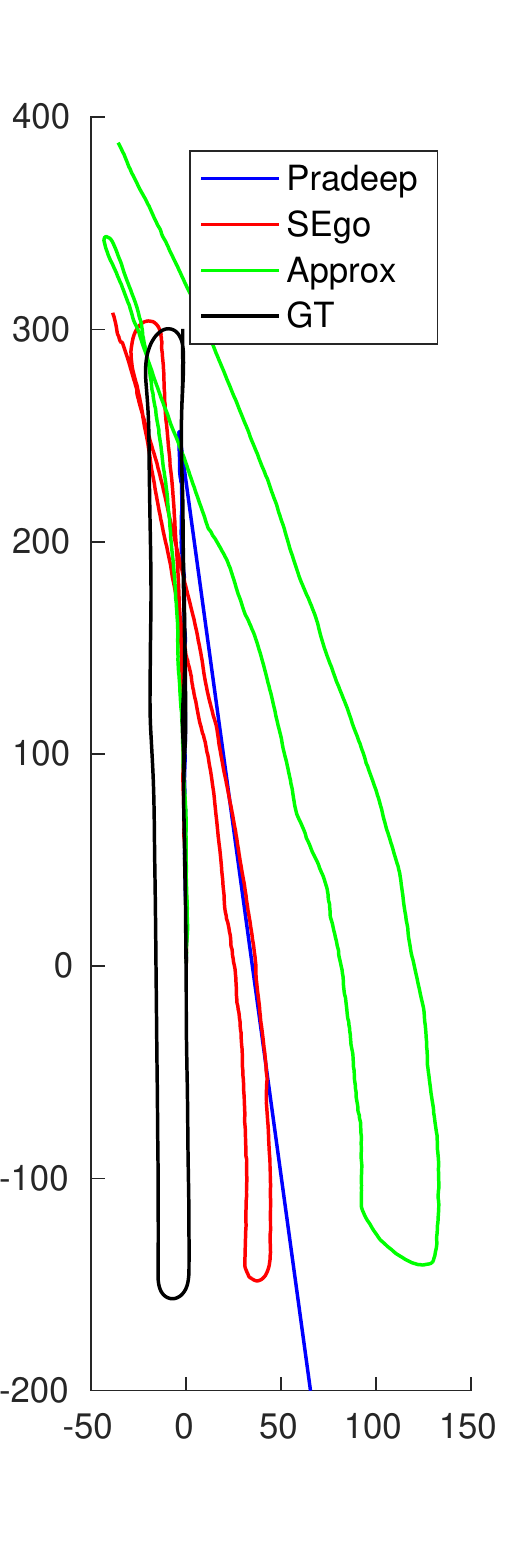}
\includegraphics[width=0.72\textwidth]{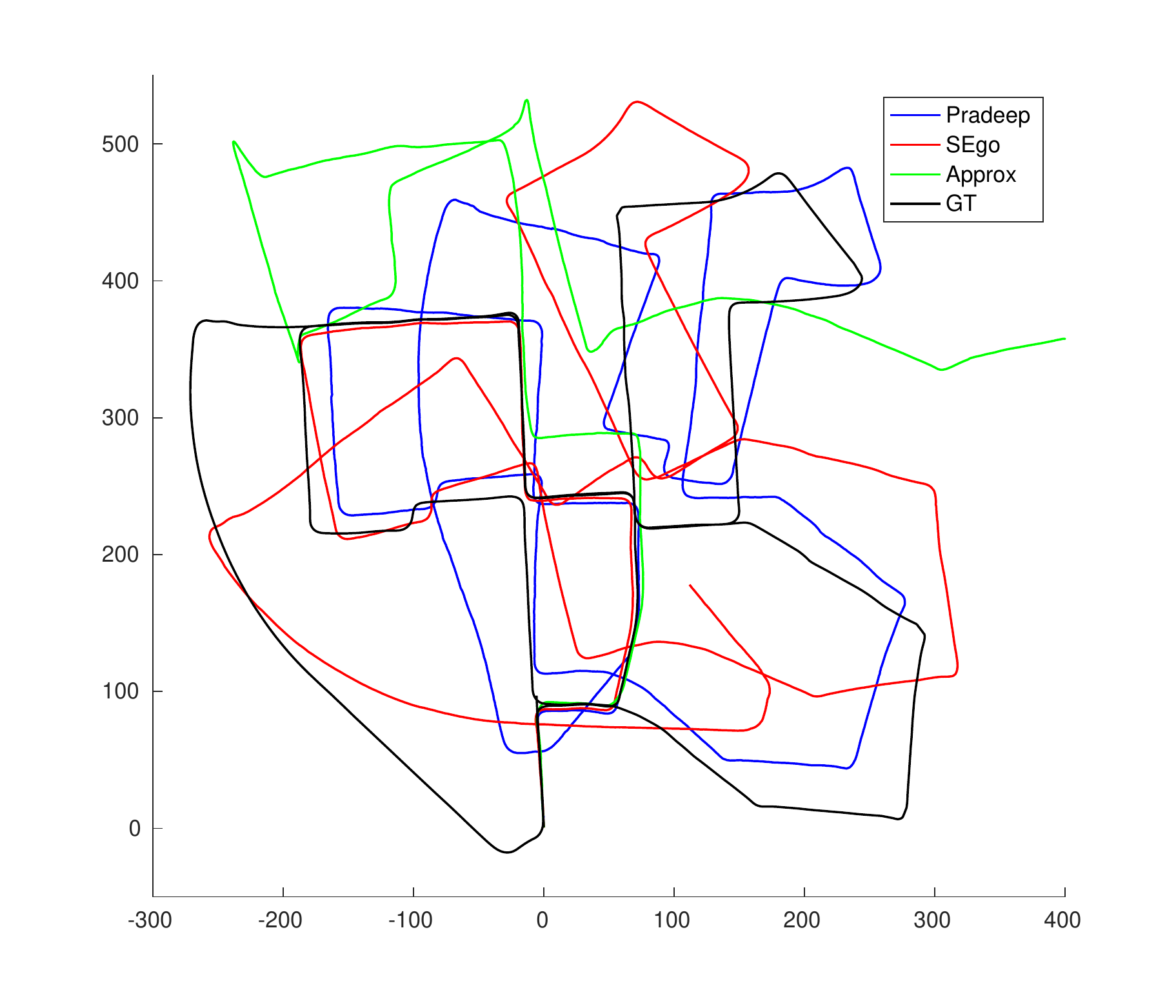}
\caption{Trajectories reconstructed by relative pose integration for the proposed solver EpiSEgo and the baselines. We use the KITTI sequences 6 (left) and 0 (right). The P3P method cannot produce a solution often so the trajectory for it is not shown. The Approx method gives low pose estimation errors compared to other baselines, but it results in large trajectory reconstruction errors due to the employed rotation approximation. The proposed solver EpiSEgo gives the closest trajectory to the ground truth.}\label{fig:trajs}
\end{figure}

\begin{figure}\label{fig:real_me_0}
	\includegraphics[width=0.5\textwidth]{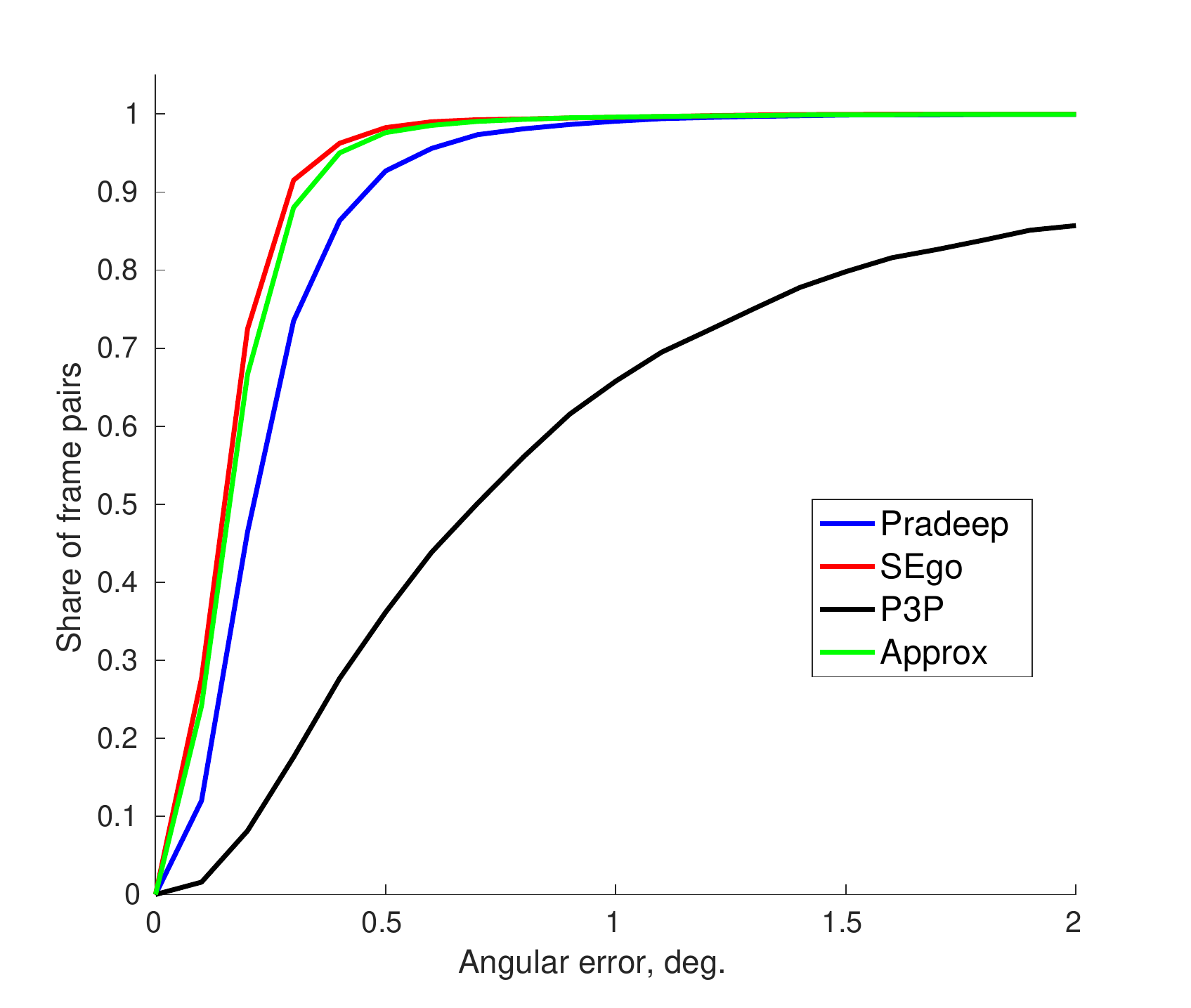}
	\includegraphics[width=0.5\textwidth]{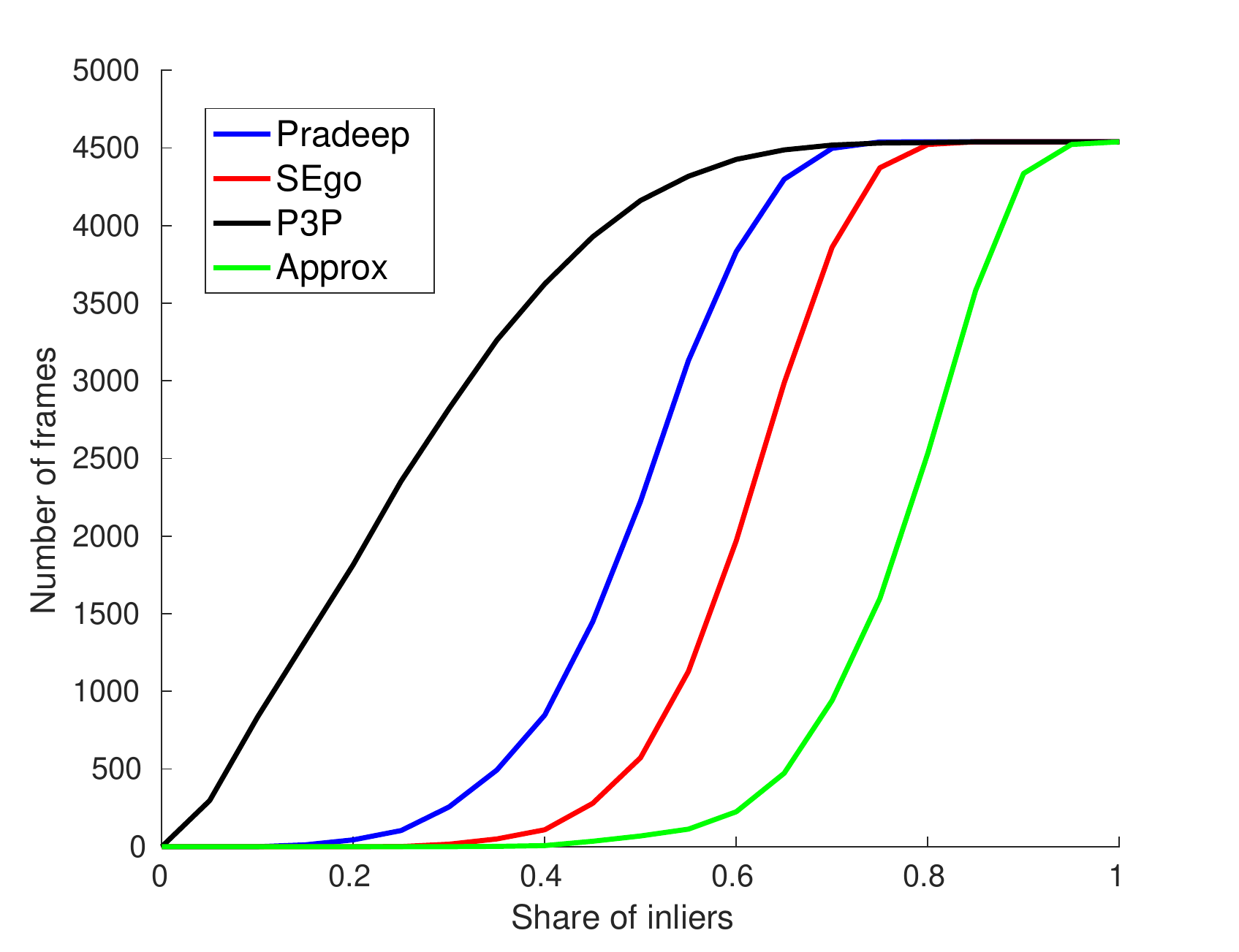}
	\caption{Results of the experiment on a KITTI sequence 0. Left: cumulative distribution for the rotation error in degrees. Right: cumulative distribution for the ratio of inliers. The proposed method EpiSEgo has higher inlier ratio than all the baselines except the Approx and lower angular error than the baselines. The Approx method has better inlier ratios due to the fact that it uses only point features.}
	\label{fig:ransaclm}
\end{figure}

\paragraph{Matching between frames.} We give here an additional illustration for the experiment with KITTI sequence 6 presented in the paper. We also show the same experiment, but with KITTI sequence 0. We reconstruct the motion trajectory by integration of the relative poses. In addition to the metrics presented in the main paper, we report the root mean-squared error (RMSE) w.r.t. the ground truth trajectory in the tab.\ref{tab:rmse}. The trajectory reconstructions for the sequences 6 and 0 are in the Fig.\ref{fig:trajs}, left and right respectively. The P3P method cannot produce a solution often so the trajectory for it is not shown. The cumulative distributions of the inlier share and angular rotation error in degrees for the KITTI sequence 0 are in the Fig.\ref{fig:real_me_0}. The Approx gives low pose estimation errors compared to other baselines, but it results in large trajectory reconstruction errors due to the employed rotation approximation. The proposed solver EpiSEgo gives the closest trajectory to the ground truth.

\paragraph{Execution time}
We report median execution times for the compared methods (see Tab.\ref{tab:time}). As we described before, the tests were performed in C++ using our (P3P, Pradeep, SEgo) and authors' (Approx) implementations. We evaluated them on a 2.3 GHz Core i5 laptop during matching between frames for the KITTI sequence 0.  

\begin{table}
\label{tab:time}
\centering
\begin{tabular}{l|c}
Method & Time, ms \\
\hline
EpiSEgo & 2 \\
Approx & 8 \\
Pradeep & 0.1 \\
P3P & 0.05\\
\end{tabular}
\caption{The median execution time (ms.) of the compared methods during matching between frames for the sequence 0. }
\end{table}

\begin{figure}
\centering
\includegraphics[width=0.22\textwidth]{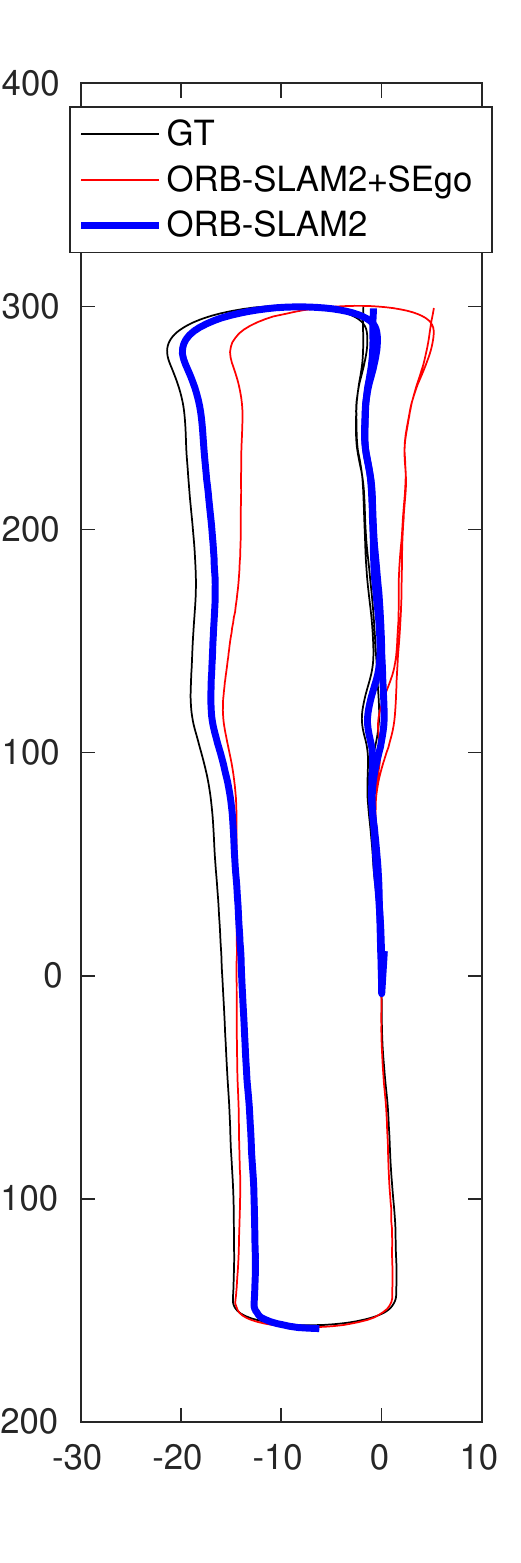}
\includegraphics[width=0.72\textwidth]{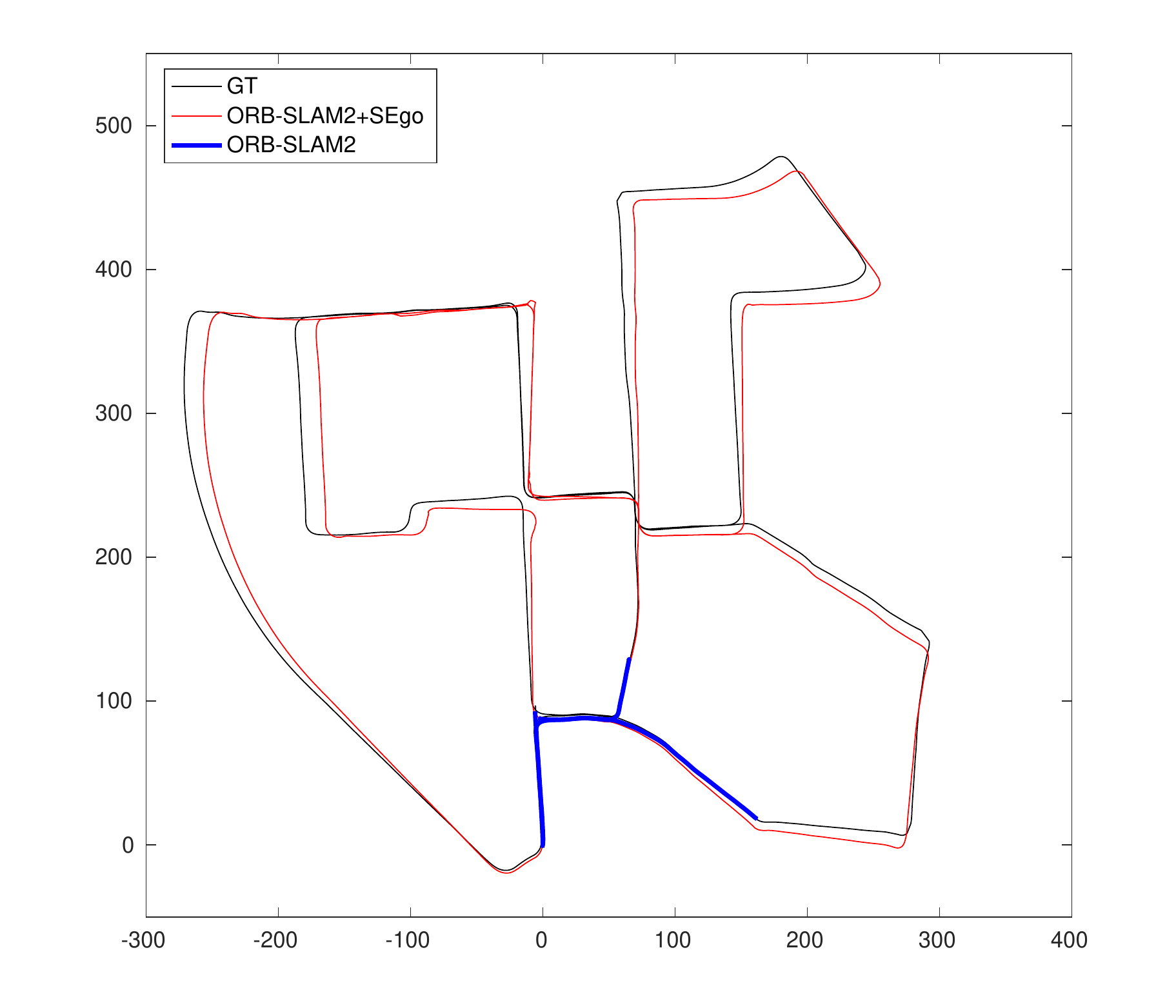}
\caption{Trajectories reconstructed by the original and modified visual SLAM pipeline \cite{mur2017orb} for the frame dropping experiment, see the paper for details. We present the trajectories for the KITTI sequences 6 (left) and 0 (right). The original pipeline loses track often. The modified pipeline tracks the full sequence. }\label{fig:trajs_orb}
\end{figure}

\paragraph{Integration into visual SLAM pipeline}
We present the (partial) trajectory reconstructions by the ORB-SLAM2 and ORB-SLAM2+SEgo pipelines for the experiments described in the paper, see fig.\ref{fig:trajs_orb}. A very slow turning motion can still be tracked by the original pipeline, but the faster motions result in track loss (see sequence 0 trajectory). The modified system can keep track even during rapid motions. 

\begin{table}
\label{tab:rmse}
\centering
\begin{tabular}{l|c|c}
Method & \multicolumn{2}{c}{$t_{RMSE}$} \\
 & 00 & 06 \\
\hline
EpiSEgo & {\bf 57.54} & {\bf 10.53} \\
Approx & 564.80 & 39.18 \\
Pradeep &76.30& 891.81 \\
P3P & - & 51457 \\
\end{tabular}
\caption{RMSE errors in meters for the trajectory reconstruction by relative pose integration for the proposed solver EpiSEgo and the baselines. We use the KITTI sequences 6 and 0 (columns). The Approx method gives low pose estimation errors compared to other baselines, but it results in large trajectory reconstruction errors due to the employed rotation approximation. The proposed solver EpiSEgo gives the closest trajectory to the ground truth.}
\end{table}

\subsection{Synthetic experiments}

\begin{table}\label{tab:code}
\centering
\begin{tabular}[width=0.5\textwidth]{l|c}
Script name & Description\\
\hline
test\_noise.m & Additive noise (Fig. 2, paper)\\ 
test\_rot.m & Rotation magnitude (Fig. 3-left, paper) \\
test\_trans.m & Translation magnitude (Fig. 4-right, paper) \\
test\_llen.m & Line length (Fig. \ref{fig:llen}, Supp. Mat.)\\
test\_noise\_planar.m & Add. noise, planar (Fig. \ref{fig:planar}, Supp. Mat.)\\
\end{tabular}
\caption{The scripts provided in the supplemental materials.}
\end{table}

\begin{figure*}
\includegraphics[width=\textwidth]{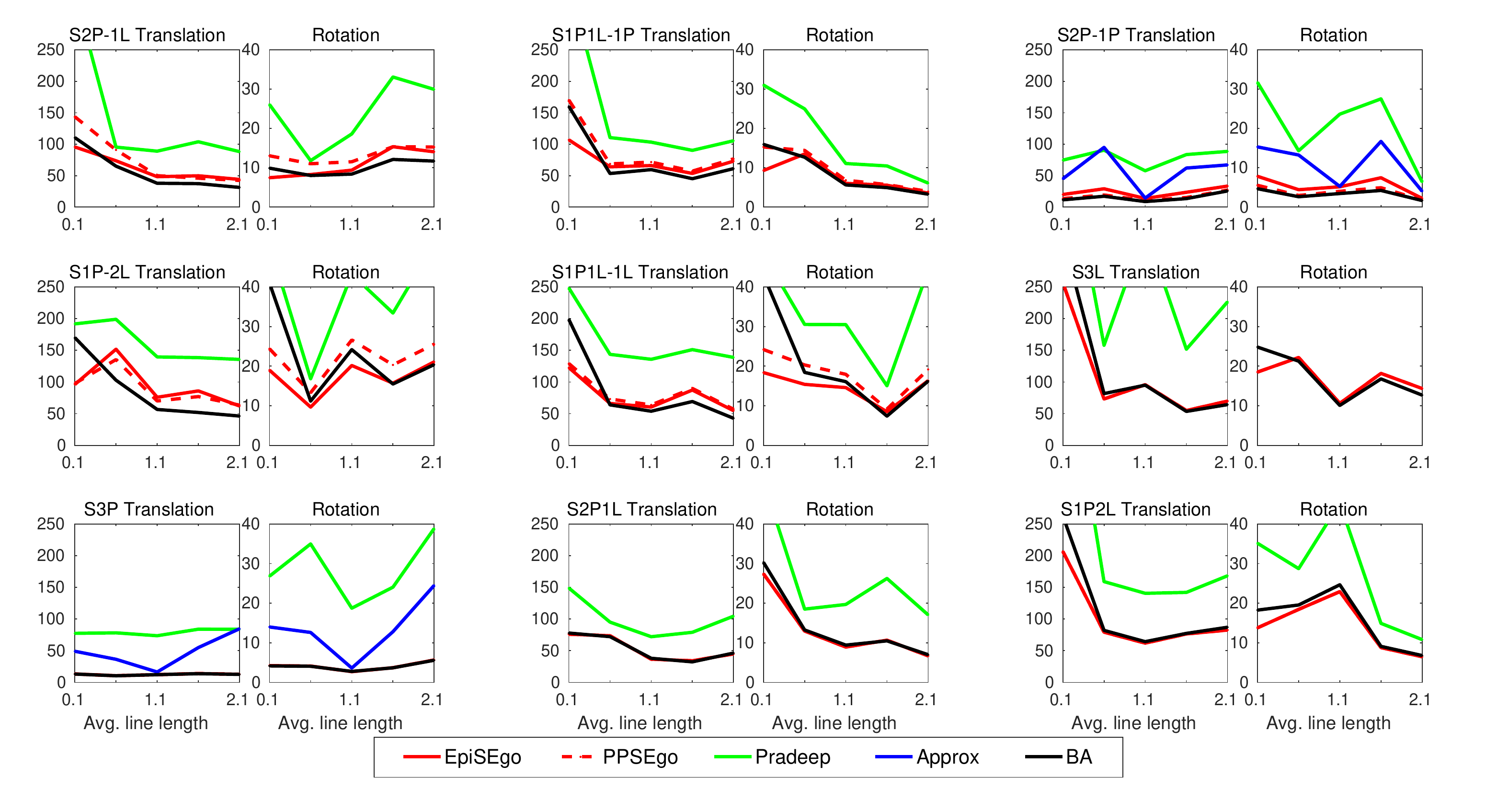}
\caption{The effect of line length variation on median relative translation and absolute rotation for each feature/correspondence combination. In the cases involving line features accuracy increases with line length. The new solvers PPSEgo and EpiSEgo are more accurate than the baselines except BA (which is given an 'unfair' advantage).  }
\label{fig:llen}
\end{figure*}

We include here the results of two more synthetic experiments. In the first one we analyze a planar case, when all the features belong to a common 3D plane. In the second one, variation of line length in the model is studied. We remind that BA is a reference method which is used to show best achievable accuracy. It is initialized with true parameter values and performs local optimization by fitting the reprojection cost to the {\it noisy} triplet feature projections. The PPSEgo and EpiSEgo both use the quadric intersection technique for the 'easy' cases. The PPSEgo uses it 'as is', while the EpiPSEgo employs the condition number check. We do so to show the benefits of using the condition number check.

\begin{figure}
\centering
\includegraphics[width=\textwidth]{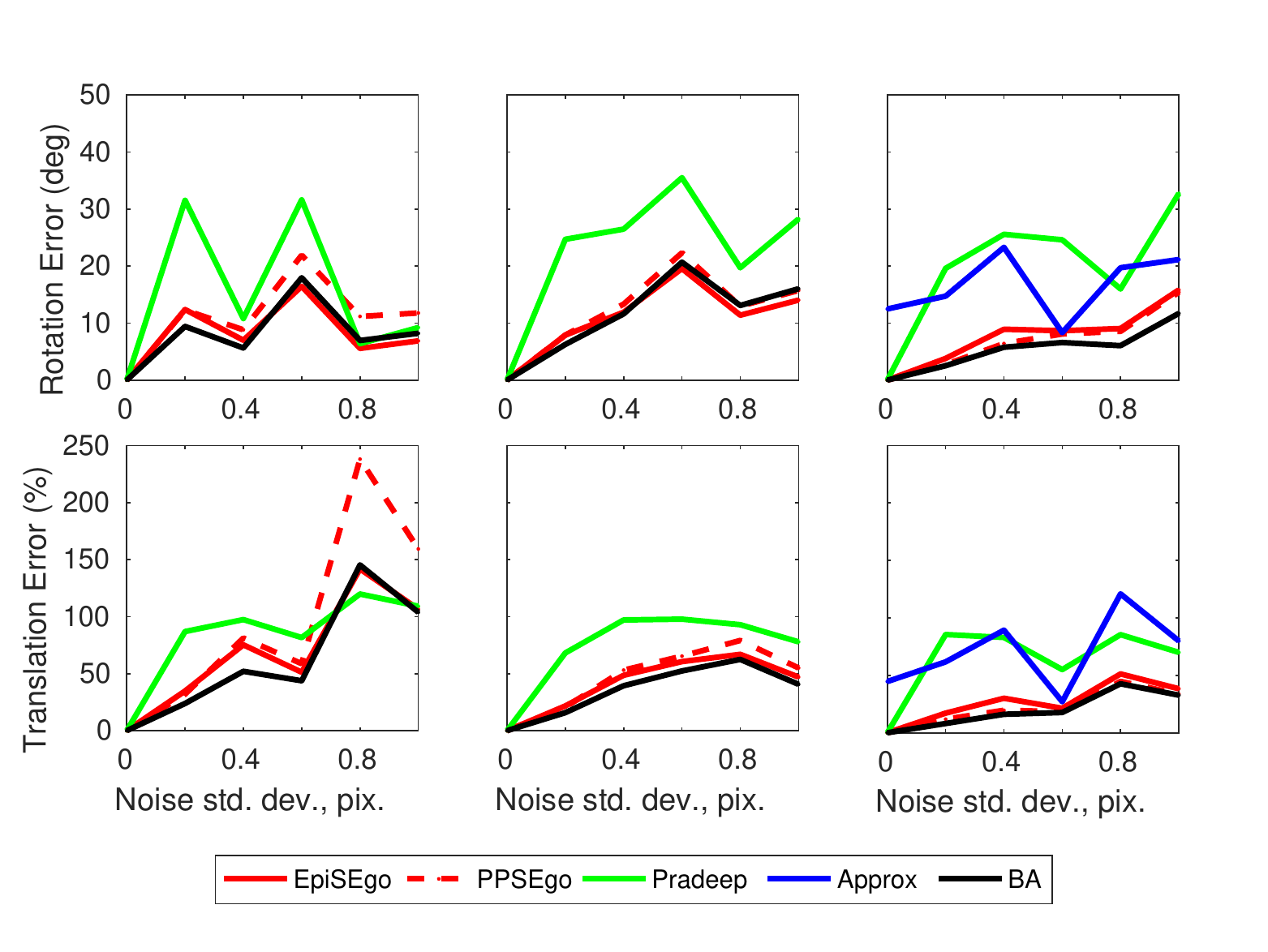}
\caption{The effect of noise magnitude variation on median relative translation and absolute rotation errors in planar case. Left graph: 'easy' cases, central graph: 'harder' cases, right graph: point-only cases.}\label{fig:planar}
\end{figure}

For the {\it planar case experiment}, we generate a random 3D plane passing through the center of the model box and project all the geometric features onto it. The other parameters of the experiment remain unchanged. We vary the noise std.dev. from 0 to 1 pix. and get the results presented in the Fig. \ref{fig:planar}. The experiment shows that the new solvers work better than the baselines with two exceptions. For the easy cases Pradeep gives lower median translation error. It uses 4-matches of features and the proposed methods use minimal 3-matches. Even the reference method BA has higher error in this case, so  it is the additional information that helps Pradeep to decrease the error. Also, in easy cases the EpiSEgo sometimes has lower accuracy than other methods. It shows importance of the condition check for the quadric intersection which is used by PPSEgo but is omitted in EpiSEgo.  

In the next experiment we vary the {\it average line length} from 0.1 to 2.0. In this case, we generate the lines in the box with given expected length. The results are in the Fig.\ref{fig:llen}. The accuracy grows with the line length for the cases in which there are line features in the minimal set. The new solvers are best in terms of accuracy except BA which has an unfair advantage being initialized with true parameter values.

\subsubsection{Code details}

We add to the supplemental materials the code reproducing our synthetic experiments in MATLAB. The scripts are described in the Tab.\ref{tab:code}. Please run mex\_setup.m  before executing them in order to compile the elimination template construction code. In order to include the Approx method into comparison, one needs to follow the instructions for C++ compilation in the README file. 

This MATLAB code is available at \url{https://github.com/alexandervakhitov/sego-paper-code.git}.
\end{document}